\documentclass{article}

\usepackage{sty/packages}
\usepackage{sty/article}
\usepackage{sty/bibliography}
\usepackage{sty/macros}

\def\tanh#1{\text{tanh} \lpar #1 \rpar}
\def\sigmoid#1{\sigma \lpar #1 \rpar}
\def\relu#1{\text{ReLU} \lpar #1 \rpar}
\def\pars#1{\left( #1 \right)}

\title{Spike-based computation using classical recurrent neural networks}

\author[1,3]{Florent De Geeter}
\author[1,2]{Damien Ernst}
\author[1]{Guillaume Drion}

\affil[1]{\normalsize Montefiore Institute, University of Liege, Liege, Belgium}
\affil[2]{\normalsize LTCI, Télécom Paris, Institut Polytechnique de Paris, France}
\affil[3]{\texttt{\normalsize florent.degeeter@uliege.be}}

\date{}

\begin{document}

\maketitle

\begin{abstract}

    Spiking neural networks are a type of artificial neural networks in which communication between
    neurons is only made of events, also called spikes. This property allows neural networks to make
    asynchronous and sparse computations and therefore drastically decrease energy consumption when
    run on specialized hardware. However, training such networks is known to be difficult, mainly due to the
    non-differentiability of the spike activation, which prevents the use of classical backpropagation. This is because
    state-of-the-art spiking neural networks are
    usually derived from biologically-inspired neuron models, to which are applied machine learning methods for training. Nowadays,
    research about spiking neural networks focuses on the design of training algorithms whose goal is to obtain
    networks that compete with their non-spiking version on specific tasks. In this paper, we attempt the symmetrical approach: we modify the dynamics of a well-known, easily trainable type of recurrent neural network to make it event-based. This new RNN cell, called the Spiking Recurrent Cell, therefore communicates using events, i.e. spikes, while being completely differentiable. Vanilla backpropagation can
    thus be used to train any network made of such RNN cell. We show that this new network can achieve performance comparable to other types of spiking networks in the MNIST benchmark and its variants, the
    Fashion-MNIST and the Neuromorphic-MNIST. Moreover, we show that this new cell makes the training of deep spiking networks achievable.

\end{abstract}

\section{Introduction}
\label{sec:introduction}

In the last decade, artificial neural networks (\textbf{ANNs}) have become increasingly powerful, overtaking human performance in many tasks.
However, the functioning of ANNs diverges strongly from the one of biological brains. Notably, ANNs require a huge amount of energy for training and inferring, whereas biological brains consumes much less power. This energy greediness prevents ANNs to be used in some environments, for instance in embedded systems.
One of the considered solutions to this problem is to replace the usual artificial neurons by spiking neurons, mimicking the function of biological brains. Spiking Neural Networks (\textbf{SNNs}) are considered as the third generation of neural networks \citep{maass_networks_1997}.
Such networks, when run on neuromorphic hardware (like \textit{Loihi} \citep{davies_loihi_2018} for instance), can show very low power consumption.
Another advantage of the SNNs is their event-driven computation.
Unlike usual ANNs that propagate information in each layer and each neuron at each forward pass, SNNs only propagate information when a spike occurs, leading to more \textit{event-driven} and sparse computations.
Nonetheless, the development of SNNs face a challenging problem: the spike generation is not differentiable, therefore preventing training using usual backpropagation \citep{rumelhart_learning_1986}, which is a the core of ANNs success.
Several solutions are being considered nowadays, as discussed in \autoref{sec:related_works}.
The classical approach consists in using a simple model for the spiking neurons to which are added learnable weights. Then, methods inspired from classical machine learning are used to train, either by directly training the SNN, for instance using surrogate gradient-descent \citep{neftci_surrogate_2019}, or by first training an ANN and then converting it into a SNN.

In this paper, we approach the problem from the other side: from the well-known \textit{Gated Recurrent Cell} (\textbf{GRU}) \citep{cho_learning_2014}, we derive a new event-based recurrent cell, called the \textit{Spiking Recurrent Cell} (\textbf{SRC}).
SRC neurons communicate via events, generated with differentiable equations.
Unlike the usual artificial spiking neurons which generate binary spikes that always last exactly one timestep, the SRC creates more natural spikes, which can span multiple timesteps.
The SRC and its equations are described in \autoref{sec:spiking_recurrent_cell}.
Such event-based cell permits to leverage the potential of classical recurrent neural networks (\textbf{RNN}) training approaches.

The performance of SRC-based RNNs has been tested on neuromorphic versions of classical benchmarks, such as the MNIST benchmark and some variants, whose results are discussed in \autoref{sec:experiments}.
To make a comparison, we have trained usual SNNs on the same benchmarks.
We show that SNNs built with SRCs achieve comparable results obtained with classic SNNs on these benchmarks for shallow networks.
But, as soon as the depth of the network increases, training becomes more and more fastidious for classic SNNs.
On the other hand, we show that SRC-SNNs composed of 10+ layers are still trainable and lead to more stable learning.

\section{Related Works}
\label{sec:related_works}

This section introduces RNNs and SNNs. Different approaches to train SNNs are also described.

\subsection{Recurrent Neural Networks}
\label{sec:recurrent_neural_networks}

RNNs are a type of neural networks that carry fading memory by propagating a vector, called the \textit{hidden state}, through time.
More precisely, a RNN is usually composed of recurrent layers, also called \textit{recurrent cells}, and classical fully-connected layers.
Each recurrent cell has its own hidden state.
At each time step, a new hidden state is computed from the received input and the previous hidden state.
This allows RNNs to process sequences.
Mathematically, this gives:
\begin{align*}
    h[t] = \phi \lpar x[t], h[t-1]; \Theta \rpar
\end{align*}
where $h[t]$ and $x[t]$ are the hidden state and the input at time $t$, respectively, $\phi$ is the recurrent cell and $\Theta$ its parameters.

Training RNNs has always been difficult, especially for long sequences, due to vanishing and exploding gradients \citep{pascanu_difficulty_2013}.
Indeed, RNNs are trained using backpropagation through time (\textbf{BPTT}) \citep{werbos_backpropagation_1990}.
This algorithm consists in first \textit{unfolding} the RNN in time, i.e. turning it into a very deep feedforward network whose number of hidden layers is equal to the sequence length and whose weights are shared among layers.
Usual backpropagation is then applied to this network.
However, due to the huge number of layers, gradient problems are much more prone to appear than in usual feedforward networks.
There exist several solutions to solve or at least attenuate these problems.
For instance, exploding gradients can be easily solved using gradient clipping \citep{pascanu_difficulty_2013}.
But the most notable improvement in RNNs was the introduction of the gating mechanism: gates, i.e. vectors of reals between 0 and 1, are used to control the flow of information, i.e. what is added to the hidden state, what is forgotten, etc.
This has led to the two most known recurrent cells: the \textit{Long-Short Term Memory} (\textbf{LSTM}) \citep{hochreiter_long_1997} and the \textit{Gated Recurrent Unit} (\textbf{GRU}) \citep{cho_learning_2014}.
LSTM uses 3 gates, while GRU is more lightweight and uses 2 gates. The new recurrent cell introduced in this paper (\autoref{sec:spiking_recurrent_cell}) is a derivation of GRU and can be expressed as an usual recurrent neural network.
\begin{new}
    Furthermore, recent work by \citet{subramoney2022Efficient}
    introduces a new type of RNN also derived from GRU, called \textit{event-based GRU}.
    This biologically inspired recurrent cell is made event-based thanks to a thresholding function, allowing the output of the cell to be only transmitted at some timesteps.
    This property allows it to be more efficient than usual RNNs, while showing competitive performance on real world tasks.
    Although it does not generate spikes and information can be contained in the event itself, it is related to our approach as both are derived from GRU and include biologically inspired mechanisms.
\end{new}

\subsection{Spiking Neural Networks}
\label{sec:spiking_neural_networks}

Biological neurons communicate using spikes, i.e. short pulses in neuron membrane potential, generated by a non-linear phenomena. These membrane potential variations are created from the flow of ions that go in and out of the cell.
There exist a lot of different mathematical models of neuron excitable membranes, the most notable being the Hodgkin-Huxley model \citep{hodgkin_quantitative_1952}, and similar models called \textit{conductance-based models}.
Such models represent the neuron membrane as a capacitance in parallel with several voltage sources and variable conductances that respectively model the electrochemical gradients of the different ions and ion membrane permeability.
Despite being very physiological, these models contain are too complex in term of variables and parameters to be used in machine learning.
That is why much more simple, phenomenological models are usually used to model spiking neurons in a SNN.

A classical model of this type is the \textit{Leaky Integrate-and-Fire} (\textbf{LIF}) model.
It is composed of a leaky integrator that integrates the input current into membrane potential variations, associated to a reset rule that is triggered when a threshold potential is reached.
Once the threshold is reached, a spike is emitted and the potential is reset to its resting value.
Unlike conductance-based models, the LIF model generates \textit{binary} spikes, i.e. spikes that last one timestep and whose value is a fixed parameter, usually set to 1 in SNNs.
Mathematically, it writes:
\begin{align*}
     & V[t] = \alpha_V V[t - 1] + x[t]
    \\
     & \begin{cases}
           \text{if } V[t] > V_{thresh} \text{, then } s[t] = 1 \text{ and } V[t] = V_{rest} \\
           \text{otherwise } s[t] = 0
       \end{cases}
\end{align*}
where $V[t]$, $x[t]$ and $s[t]$ are the membrane potential, the input and the output at time $t$, respectively, $\alpha_V$ is the leakage factor, $V_{thresh}$ the threshold and $V_{rest}$ the resting potential. The LIF model is far less physiological then conductance-based models, but it is much more lightweight and retains the core of spike-based computation.

LIF neurons can be organized in layers to form a complete network.
The question is now how to train such a network?
Due to the non-differentiable activation, usual backpropagation cannot be used (or at least cannot be used directly).
To achieve reasonable training performance, many approaches to train SNNs have been proposed \citep{yamazaki_spiking_2022,tavanaei_deep_2018}, which can be split into three categories.
First, SNNs can be trained using unsupervised learning rules, which are local to the synapses \citep{masquelier_unsupervised_2007,neftci_event-driven_2014,diehl_unsupervised_2015,lee_deep_2019}.
These learning rules are often derived from the \textit{Spike-timing-dependent plasticity} process \citep{markram_regulation_1997}, which strengthens or weakens synaptic connections depending on the coincidence of pre and post-synaptic spikes.
This non-optimization-based training method is usually slow, often unreliable, and leads to poor performance.
The second category is an indirect training.
It consists in first training a usual ANN (with some constraints) and then converting it into a SNN \citep{cao_spiking_2015,diehl_fast-classifying_2015,esser_convolutional_2016}.
Indeed, ANNs can be seen as special spiking networks that uses a rate-based coding scheme.
These methods allow to use all the algorithms developed for training ANNs, and thus can reach high performance.
However, they do not unlock the full potential of spiking networks, as rate-coding is not the only way of transmitting information through spikes.
Also, rate-based coding usually results in a higher number of generated spikes, weakening the energy-efficiency of SNNs.
The third and last approach is to rely on gradient-based optimization to directly train the SNN \citep{bohte_spikeprop_2000,sporea_supervised_2013,hunsberger_spiking_2015,zenke_superspike_2018,shrestha_slayer_2018,lee_training_2016,neftci_surrogate_2019}.
These methods usually smooth the entire networks or use a surrogate smoothed gradient for the non-differentiable activation to allow backpropagation.
Notably, \citet{huh_gradient_2018} used a smooth spike-generating process which replaces the non-differentiable activation of the LIF neurons.
This approach is closely related to ours, as they both use soft non-linear activations to generate spikes.
SNNs trained by gradient-based algorithms have achieved good performance, even competing with ANNs on some benchmarks.
However, it is known to be very difficult to train deep SNNs.
Indeed, adding more layers makes the training much harder, which prevents from obtaining efficient deep SNNs and therefore limits the complexity of the tasks SNNs can solve.

\section{Spiking Recurrent Cell}
\label{sec:spiking_recurrent_cell}

The new spiking neuron introduced in this paper is derived from the well-known recurrent neural network GRU.
This section describes its derivation and the different parts of the neuron, namely the spike-generation and the input-integration parts.

\subsection{Spike-generation}
\label{sec:spike_generation}

As the starting point of the derivation of the SRC equations, another recurrent cell will be used, itself derived from GRU: the \textit{Bistable Recurrent Cell} (\textbf{BRC}) created by \citet{vecoven_bio-inspired_2021}.
Its main property is its never-fading memory created by the bistability property of its neurons.

Here are the equations of GRU:
\begin{subequations}
    \begin{align}
        z[t]       & = \sigmoid{U_z x[t] + {W_z} h[t-1] + b_z} \label{eq:gru_z}
        \\
        r[t]       & = {\sigma} \pars{U_r x[t] + {W_r} h[t-1] + b_r} \label{eq:gru_r}
        \\
        \hat{h}[t] & = \tanh{U_h x[t] + r[t] \odot {W_h} h[t-1] + b_h} \label{eq:gru_cand}
        \\
        h[t]       & = z[t] \odot h[t-1] + \lpar 1 - z[t] \rpar \odot \hat{h}[t] \label{eq:gru_h}
    \end{align}
\end{subequations}

And here are the ones of BRC:
\begin{subequations}
    \begin{align}
        z[t]       & = \sigma \lpar U_z x[t] + {w_z \odot} h[t-1] + b_z \rpar \label{eq:brc_z}
        \\
        r[t]       & = {1 + \text{tanh}} \lpar U_r x[t] + {w_r \odot} h[t-1] + b_r \rpar \label{eq:brc_r}
        \\
        \hat{h}[t] & = \tanh{U_h x[t] + r[t] \odot h[t-1] + b_h} \label{eq:brc_cand}
        \\
        h[t]       & = z[t] \odot h[t-1] + \lpar 1 - z[t] \rpar \odot \hat{h}[t] \label{eq:brc_h}
    \end{align}
\end{subequations}

Both use two gates ($z$ and $r$) to control the flow of information.
$r$ is used to compute what we will call the \textit{candidate} hidden state $\hat{h}[t]$.
$r$ controls how much of the previous hidden state will be included in $\hat{h}[t]$.
$z$ is then used to compute the next hidden state from the previous one and the candidate one.
It controls the ratio of history and novelty in the new hidden state.

There are two major differences between GRU and BRC.
First, the memory in BRC is cellular, meaning that each neuron of the cell has its own internal memory that is not shared with the others, while in GRU all internal states can be accessed by each neuron.
This is constrained in the equations by the transformation of the matrices $W_z$ and $W_r$ (\autoref{eq:gru_z} and \autoref{eq:gru_r}) into the vectors $w_z$ and $w_r$ (\autoref{eq:brc_z} and \autoref{eq:brc_r}) as well as the removal of $W_h$ (\autoref{eq:gru_cand}) in \autoref{eq:brc_cand}.
The second difference is the range of possible values of $r$: in GRU, it is included between $0$ and $1$ while in BRC, it is included between $0$ and $2$.
This difference allows the BRC neuron to switch from monostability ($r \leq 1$) to bistability ($r > 1$).
This is reflected in the equations by a change in the activation of $r$: GRU uses a sigmoid (\autoref{eq:gru_r}), while BRC uses a shifted hyperbolic tangent (\autoref{eq:brc_r}).

These two properties of BRC, i.e. the cellular memory and the bistability, are useful to generate spikes.
The cellular memory can represent the membrane potential of the spiking neurons, while the bistability is created by a local positive feedback, which is the first step of a spike.
Indeed, a spike can be described in two steps: a fast local positive feedback that brings the potential to a high value followed by a slower global negative feedback that brings back the potential to its resting value.
Therefore, integrating such a negative feedback to BRC equations allows the cell to generate spikes.
This can be done by adding a second hidden state $h_s$, which lags behind $h$ (\autoref{eq:h_slow}), and a new term in the computation of $\hat{h}$ (\autoref{eq:h_hat}).
As no information can be transmitted between neurons except when a spike occurs, the fast hidden state $h$ is passed through a ReLU function to isolate the spikes from the small, subthreshold variations of $h$.
This creates the output spikes train $s_{out}$ (\autoref{eq:y}).
The input of SRC, i.e. the integration of the input pulses, will be discussed afterwards, therefore we will simply use $x$ to denote the input used by the spike generation.

This leads to the equations that generate spikes:
\begin{subequations}
    \begin{align}
        \hat{h}[t] & = \tanh{x[t] + r \odot h[t-1] \, {+ \, r_s \odot h_s[t-1]} + b_h} \label{eq:h_hat}
        \\
        h[t]       & = z \odot h[t-1] + \pars{1 - z} \odot \hat{h}[t] \label{eq:h_fast}
        \\
        h_s[t]     & = z_s \odot h_s[t-1] + \pars{1 - z_s} \odot h[t-1] \label{eq:h_slow}
        \\
        s_{out}[t] & = \relu{h[t]} \label{eq:y}
    \end{align}
\end{subequations}

From the point of view of RNNs, these equations contain four gates: $r$, $r_s$, $z$ and $z_s$.
$r$ and $r_s$ control the impact of the previous hidden states $h[t-1]$ and $h_s[t-1]$ on the candidate hidden state, while $z$ and $z_s$ control the update speed of $h$ and $h_s$.
To understand how these equations allow to generate spikes, we should approach them from the perspective of dynamical systems.
Indeed, spike creation comes from a mix of feedbacks: first a positive one, followed by a negative one.
These feedbacks are initiated through the recurrent connections of \autoref{eq:h_hat}, and the coefficients $r$ and $r_s$ control the feedbacks strengths.
The fast feedback must be generated by the fast hidden state, i.e. $h$, therefore its coefficient $r$ should be positive and high enough.
Following the same logic, the slow feedback is brought by $h_s$.
Its coefficient $r_s$ should thus be negative.
The control of the timescales, i.e. the convergence speed of $h$ and $h_s$, is managed by the two other coefficients, $z$ and $z_s$ along with \autoref{eq:h_fast} and \autoref{eq:h_slow}.
$h$ must be very fast, thus $z$ should be close to $0$, while $h_s$ must lag behind $h$, imposing $z_s$ to have a greater value.
To enforce that no computation could be achieved through alterations in the shape of a spike, the 4 gates cannot depend anymore on learnable weights.
We have fixed three of them with constant values: $r = 2$, $r_s = -7$ and $z = 0$.
The last one, $z_s$, which controls the convergence speed of $h_s$ towards $h$, has been made dependent of $h$.
To create spikes with short depolarization periods, $z_s$ should be low at depolarization potentials (high $h$), and larger at subthreshold potentials (low $h$), mimicking the voltage-dependency of ion channel time constants in biological neurons.
Therefore, we have decided to use a simple step function to allow $z_s$ to jump between two values:
\begin{align*}
    z_s[h] & = z_s^{hyp} + (z_s^{dep} - z_s^{hyp}) \times H \lpar h[t] - 0.5 \rpar
\end{align*}
where $z_s^{hyp}$ is the value of $z_s$ when $h$ is low, $z_s^{dep}$ the value of $z_s$ when $h$ is high and $H$ denotes the Heaviside step function.
In practice, we use $z_s^{hyp} = 0.9$ and $z_s^{dep} = 0$.

Finally, the bias $b_h$ controls the propensity of neurons to fire spikes: the higher, the easier.
However if it reaches too high values, the neurons may saturate.
As this is a behavior that we would rather avoid, the bias should be constrained to always be smaller than some value.
In the experiments, we have fixed this higher bound to $-4$.

\begin{figure}
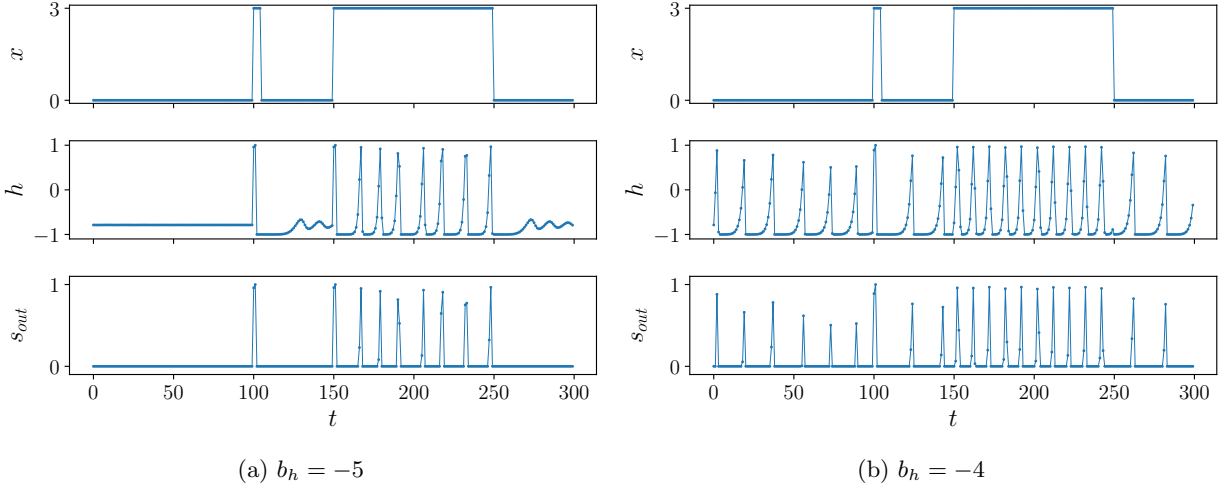

    \centering
    \begin{subfigure}{.49\textwidth}
        \centering
        \includegraphics[width=\textwidth]{assets/pdf/src_dynamics/src_bias5.pdf}
        \caption{$b_h = -5$}
        \label{fig:src_dynamics_bias-5.0}
    \end{subfigure}
    \begin{subfigure}{.49\textwidth}
        \centering
        \includegraphics[width=\textwidth]{assets/pdf/src_dynamics/src_bias4.pdf}
        \caption{$b_h = -4$}
        \label{fig:src_dynamics_bias-4.0}
    \end{subfigure}
    \caption{Simulation of a SRC neuron for some inputs sequence $x$ and different biases $b_h$.}
    \label{fig:src_dynamics}
\end{figure}

\autoref{fig:src_dynamics} shows the behavior of one SRC neuron given different inputs $x$ and biases $b_h$.
It can be observed that for a high bias (\autoref{fig:src_dynamics_bias-4.0}), the neuron is able to spike even with a null input, while for a lower one (\autoref{fig:src_dynamics_bias-5.0}), the neuron remains silent.
\autoref{sec:feedback_dynamics_in_src} explains in more details the spike generation of SRC by showing the interaction between the negative and positive feedbacks.

SNNs are often put forward for their very small energy consumption, due to the sparse activity of spiking neurons.
It is thus important to be able to measure the activity of said neurons.
In the context of SRC neurons, the spikes do not last exactly one timestep.
It is therefore better to compute the number of timesteps during which spikes are emitted rather than the number of spikes.
This brings us to define the relative number of \textit{spiking} timesteps:
\begin{align}
    \mathcal{T}(s) = \frac{1}{T} \sum_{t = 1}^{T} H(s[t]),
    \label{eq:spiking_time}
\end{align}
where $T$ is the number of timesteps.

\subsection{Input-integration}
\label{sec:inputs_integration}

The last point to be addressed before being able to construct networks of SRCs is how to integrate the input spikes $s_{in}$.
We decided to use leaky integrators with learnable weights $w_i$:
\begin{align*}
    i[t] = \alpha \, i[t-1] + \sum_i w_i s_{in}[t]
\end{align*}
where $\alpha$ is the leakage factor.

To prevent the SRC from saturating due to large inputs, we also added a rescaled hyperbolic tangent to $i$ to create neuron input $x$. The rescaling factor $\rho$ is set to $3$, forcing $x$ to be between $-3$ and $3$.

The equations of a whole SRC layer therefore writes, starting from the input pulses $s_{in}$ up to the output pulses $s_{out}$:
\begin{subequations}
    \begin{align}
        i[t]       & = \alpha \, i[t-1] + W_s \, s_{in}[t] \label{eq:src_i}
        \\
        x[t]       & = \rho \cdot \tanh{\frac{i[t]}{\rho}} \label{eq:src_x}
        \\
        z_s[h]     & = z_s^{hyp} + (z_s^{dep} - z_s^{hyp}) \times H \lpar h[t] - 0.5 \rpar \label{eq:src_zs}
        \\
        h[t]       & = \tanh{x[t] + r \odot h[t-1] + r_s \odot h_s[t-1] + b_h} \label{eq:src_h}
        \\
        h_s[t]     & = z_s[h] \odot h_s[t-1] + \pars{1 - z_s[h]} \odot h[t-1] \label{eq:src_hs}
        \\
        s_{out}[t] & = \relu{h[t]} \label{eq:src_s}
    \end{align}
\end{subequations}

To sum up, \autoref{eq:src_i} first integrates the input pulses using a leaky integrator.
The result then passes through a rescaled hyperbolic tangent in \autoref{eq:src_x}.
$z_s$ is computed, based on $h$, in \autoref{eq:src_zs}.
This forms the input used by the \textit{spike generation} part (\autoref{eq:src_h} and \autoref{eq:src_hs}) to update $h$ and $h_s$.
Finally, \autoref{eq:src_s} isolates the spikes from the small variations of $h$ and generates the output pulses. Finally, like the other recurrent cells, SRC can be organized in networks with several layers.

\begin{new}
\begin{table}
    \centering
    \begin{tabular}{|l|l|c|}
        \hline
        \textbf{Name} & \textbf{Description} & \textbf{Typical value} \\
        \hline
        $\alpha$ & Leakage factor of the integrator & $0.9$/$0.99$ \\
        $W_s$ & Learning weights & \\
        $\rho$ & Rescaling factor to avoid saturation & $3.$ \\
        $r$ & Positive feedback factor & $2.$ \\
        $r_s$ & Negative feedback factor & $-7.$ \\
        $b_h$ & Bias of $h$, controls the propensity to spike & $-6.$ \\
        $z$ & Controls the convergence speed of $h$ towards $\hat{h}$ & $0.$ \\
        $z_s^{hyp}$ & Controls the convergence speed of $h_s$ towards $h$ at low values of $h$ & $0.9$ \\
        $z_s^{dep}$ & Controls the convergence speed of $h_s$ towards $h$ at high values of $h$ & $0.$ \\
        \hline
    \end{tabular}
    \caption{Summary of the parameters present in the equations of SRC.}
    \label{tab:src_parameters}
\end{table}

SRC has indeed a non negligible number of parameters, especially compared to simple models like LIF.
However, these are required to create an artificial spiking neuron able to mimic, to some extent, the subthreshold dynamics of real neurons.
Most of these parameters should not be learned, as they ensure fixed and stable dynamics to generate the spikes and to avoid saturation of the cell.
\autoref{tab:src_parameters} summarizes the parameters of SRC and briefly describes them.
\end{new}

\section{Experiments}
\label{sec:experiments}

This section describes the different experiments that were made to assess SRC performance. We have used PyTorch to implement the cell and to train it. To make a comparison, we have trained LIF SNNs on the same benchmarks. To train them, we have used the snnTorch library \citep{eshraghian2021training}, which allows to easily train SNNs using the surrogate gradient approach and is based on PyTorch. All the parameters are given in \autoref{sec:experiments_details}.

\subsection{Benchmarks}
\label{sec:benchmarks}
The SRC has been tested on the well-known MNIST dataset \citep{deng_mnist_2012}, as well as two variants.
The first variant is the Fashion-MNIST dataset \citep{xiao2017fashionmnist}, that contains images of fashion products instead of handwritten digits.
It is known to be more difficult than the original MNIST.
The second variant is the Neuromorphic MNIST (N-MNIST) \citep{orchard_converting_2015} which, as its name suggests, is a neuromorphic version of MNIST where the handwritten digits have been recorded by an event-based camera.

The MNIST and Fashion-MNIST datasets are not made to be used with spike-based networks, therefore their images must first be encoded into spike trains.
To do so, a rate-based coding and a latency-based coding were used in the experiments.
The first one creates one spike train per pixel, where the number of spikes per time period is proportional to the value of the pixel.
More precisely, the pixel is converted into a Poisson spike train using its value as the mean of a binomial distribution.
To avoid having too many spikes, we have scaled the pixel values by a factor (the \textit{gain}) of $0.25$.
Therefore, a white pixel (value of $1$) will spike with a probability of 25\% at each timestep, while a black one (value of $0$) will never spike.
The latency-based coding is much more sparse, as each pixel will spike at most one time.
In this case, the information is contained in the time at which the spike occurs.
The idea is that brighter pixels will spike sooner than darker ones.
The spike time $t_{spk}$ of a pixel is defined as the duration needed by the potential of a (linearized) RC circuit to reach a threshold $V_{th}$ if this circuit is alimented by a current $I$ equivalent to the pixel value:
\begin{align*}
    t_{spk} = \text{min} \lpar - \tau \lpar I - 1 \rpar, V_{th} \rpar
\end{align*}
where $\tau$ is the time constant of the RC circuit.
In our experiments, we have used a $\tau = 10$ and a $V_{th} = 0.01$.
The spike times are then normalized to span the whole sequence length, and the spikes located at the last timestep (i.e. the spikes whose $t$ equals to $V_{th}$) are removed.
The encodings were performed using the snnTorch library \citep{eshraghian2021training}.
All the experiments were made using spikes trains of length 200.
Therefore, the MNIST (or Fashion-MNIST) inputs of dimension $(1, 28, 28)$ are converted to tensors of size $(200, 1, 28, 28)$.

On the other hand, N-MNIST already has event-based inputs.
Indeed, each sample contains the data created by an event-based camera.
Therefore this data just need to be converted to tensors of spikes.
An event-based camera pixel outputs a event each time its brightness changes.
There are therefore two types of events: the ones issued when the brightness increased and the ones issued when it decreases.
A N-MNIST sample is a list of such events, which contains a timestamp, the coordinates of the pixel that emitted it, and its type.
The Tonic library \citep{lenz_gregor_2021_5079802} was used to load the N-MNIST dataset and convert its samples into tensors of size $(200, 2, 34, 34)$.
The first dimension is the time, the second is related to the type of the event and the two last are the x and y spatial coordinates.

\subsection{Readout layer}
\label{sec:readout_layer}

In order to extract the predictions from the outputs of a SNN, the final layer is connected with predefined and frozen weights to a readout layer of leaky integrators, with one integrator per label and a leakage factor of $0.99$. Each integrator is excited (positive weight) by a small group of neurons and is inhibited (negative weight) by the others.
In our experiments, this final layer contains 100 neurons.
Each integrator is connected to all neurons: 10 of these connections have a weight of 1, while the others have a weight of -0.1.
The prediction of the model corresponds to the integrator with the highest value at the final timestep.

\subsection{Loss function}
The networks were trained using the cross-entropy loss, which is usually used in classification tasks. This function takes as inputs the values $x$ of the leaky integrators at the final timestep and the target class $y$.
The loss is then computed (for a single sample) as:
\begin{align*}
    l(x, y) & = -log \lpar \frac{exp(x_y)}{\sum_{c = 1}^{C} exp(x_c)}  \rpar
\end{align*}
where $C$ is the number of classes and $x_c$ refers to the element of $x$ associated to the class $c$.

This function basically applies the \textit{Softmax} function to $x$ and then computes the negative log likelihood.
For a whole batch, we simply take the mean of the $l$'s.

\subsection{Learning}

The loss function being defined, it is now possible to train networks of SRCs using the usual automatic differentiation of PyTorch.

SRCs contain more parameters than LIFs, however most of them are not learnable, i.e. their values are fixed at initialization and they are not updated during training.
In addition to the weights $W_s$, only the bias $b_h$ has been made learnable.
While it is possible to make other parameters learnable, we wanted to focus on the network ability to learn through changes in weights and biases.

Before presenting the results, there are two last things that must be discussed.
The first point concerns the backpropagation through the layers.
Experiments showed that bypassing the ReLU during backpropagation improves learning.
As explained in \autoref{sec:spike_generation}, this ReLU is used to isolate the spikes (high variations of $h$) from the small fluctuations.
Considering the backward pass, this ReLU \textit{blocks} the gradients when no spike is currently occurring, i.e. $h[t] < 0$.
We therefore tested to let these gradients pass even when no spike is occurring:
\begin{align*}
     & s_{out}[t] = \relu{h[t]}                                       \\
     & \frac{\partial s_{out}[t]}{\partial h[t]} = 1, \, \forall h[t]
\end{align*}

\begin{figure}
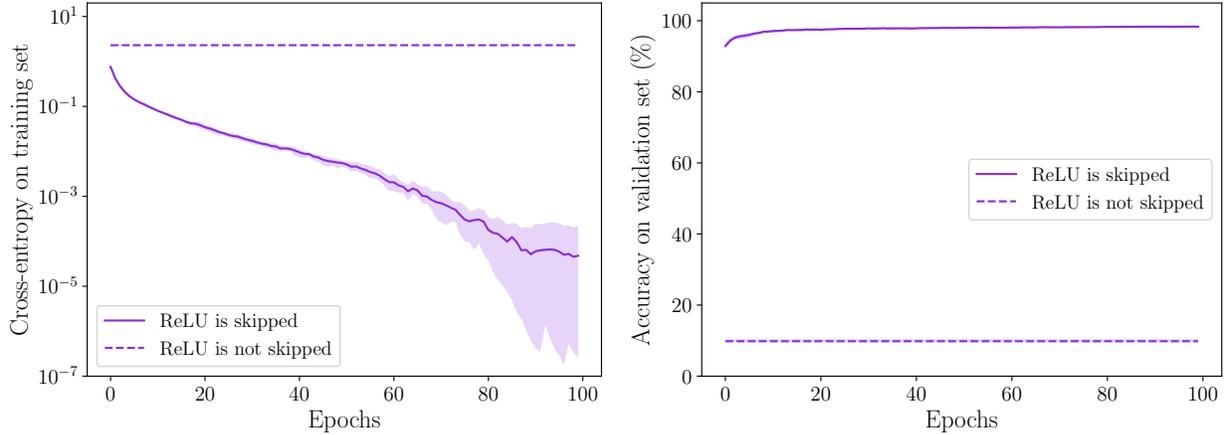

    \centering
    \begin{subfigure}{\textwidth}
        \centering
        \includegraphics[width=.49\textwidth]{assets/pdf/from_wandb/skip_relu_ce.pdf}
        \includegraphics[width=.49\textwidth]{assets/pdf/from_wandb/skip_relu_acc.pdf}
    \end{subfigure}
    \caption{Evolution of the training cross-entropy and the validating accuracy on the MNIST dataset for two sets of networks composed of 5 SRC layers. For one set, the ReLU has been skipped during the backpropagation.}
    \label{fig:surrogate}
\end{figure}

\autoref{fig:surrogate} shows the evolution of the accuracy and cross-entropy of two SRC networks composed of 5 layers, one trained with the ReLU bypass, the other without.
For each network we have trained 5 models.
Except the bypass of the ReLU, all the other parameters are the same.
When the gradients are back-propagated through the ReLU, the models do not manage to learn, while when the ReLU is skipped, learning is achieved, fast and leads to a good performance.
We have therefore decided to use this trick for all the experiments.
Although, note that the ReLU is still used during forward passes, to ensure that information is transmitted between neurons only when spikes occur.

The second point concerns the backpropagation through time.
A SRC neuron has three states: $i$, $h$ and $h_s$.
This means that gradients can be backpropagated through time via three different \textit{paths}.
However, we found out empirically that preventing the gradients from flowing though the recurrent connections of $h$ and $h_s$ improves learning.
In this case, only the state of the leaky integrator, hence the synapse, is used to transmit gradients back in time, which is similar to what is done in LIF neurons.
The improvement is not as important as skipping the ReLU, yet we decided to use it in the experiments.
In practice, this can be easily done in PyTorch, using the \textit{detach} method.

\subsection{Results}
\label{sec:results}

This subsection describes the results obtained from the different experiments we made with SRC and LIF SNNs.
First, shallow networks have been tested on the different benchmarks to compare the performances of both neurons.
Then, deeper networks have been tested to analyse the capacity of learning in depth.
All the parameters are given in \autoref{sec:experiments_details}.

\subsubsection{Shallow networks}

As a first experiment, we have tested several shallow networks with either 1, 2, 3 or 5 layers on the different benchmarks.

\begin{table}[!ht]
    \centering
    \begin{tabular}{|l|l|c|c|c|c|c|c|c|c|c|c|}
        \hline
        \multirow{2}{3.em}{Layers} & \multirow{2}{3.2em}{Neuron} & \multicolumn{2}{c|}{MNIST} & \multicolumn{2}{c|}{Fashion MNIST} & \multirow{2}{4.4em}{N-MNIST}                   \\
        \cline{3-6}
                                   &                             & Rate                       & Latency                            & Rate                         & Latency &       \\
        \hline
        \multirow{2}{2.7em}{1}     & SRC                         & 97.27                      & 96.54                              & 86.24                        & 86.28   & 97.42 \\
        \cline{2-7}
                                   & LIF                         & 97.18                      & 96.98                              & 86.02                        & 84.72   & 98.18 \\
        \hline
        \multirow{2}{2.7em}{2}     & SRC                         & 98.32                      & 98.30                              & 87.80                        & 87.06   & 98.25 \\
        \cline{2-7}
                                   & LIF                         & 98.06                      & 96.92                              & 79.32                        & 87.27   & 97.34 \\
        \hline
        \multirow{2}{2.7em}{3}     & SRC                         & 98.44                      & 98.21                              & 88.35                        & 84.86   & 98.34 \\
        \cline{2-7}
                                   & LIF                         & 98.18                      & 94.73                              & 80.64                        & 78.09   & 97.36 \\
        \hline
        \multirow{2}{2.7em}{5}     & SRC                         & 98.39                      & 97.52                              & 88.54                        & 82.09   & 98.11 \\
        \cline{2-7}
                                   & LIF                         & 98.08                      & 97.61                              & 84.22                        & 85.88   & 98.05 \\
        \hline
    \end{tabular}
    \caption{Mean accuracies (in \%) obtained on the test sets of the different datasets and the different encodings.}
    \label{tab:results}
\end{table}

\begin{figure}[htb]
    \centering
    \includegraphics[width=.49\textwidth]{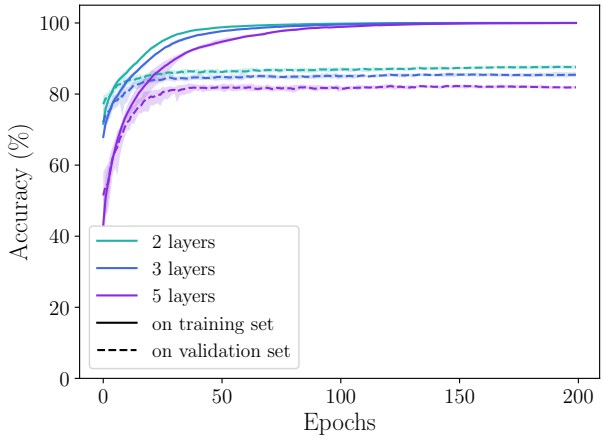}
    \caption{Evolution of the training and validating accuracies of SRC networks with 2, 3 and 5 layers on the Fashion MNIST with latency-based coding.}
    \label{fig:overfitting}
\end{figure}

\autoref{tab:results} shows the different testing accuracies achieved by these networks.
We can observe that SRC networks were able to learn and achieved equal and sometimes superior performances compared to LIF networks.
Also, adding hidden layers usually leads to better performance.
However, we noticed that SRC networks were prone to overfitting in some cases.
For instance, when trained on the Fashion MNIST with a latency-based coding, their performance decreases when hidden layers are added.
\autoref{fig:overfitting} shows the evolution of the accuracies on training and validation sets for the SRC networks.
All networks with hidden layers manage to reach a training accuracy very close to $100\%$, but the validating accuracy obtained at the end decreases with the number of hidden layers.
This problem of overfitting it is well-known in classic deep learning, and further research might find ways to reduce it.

\begin{figure}
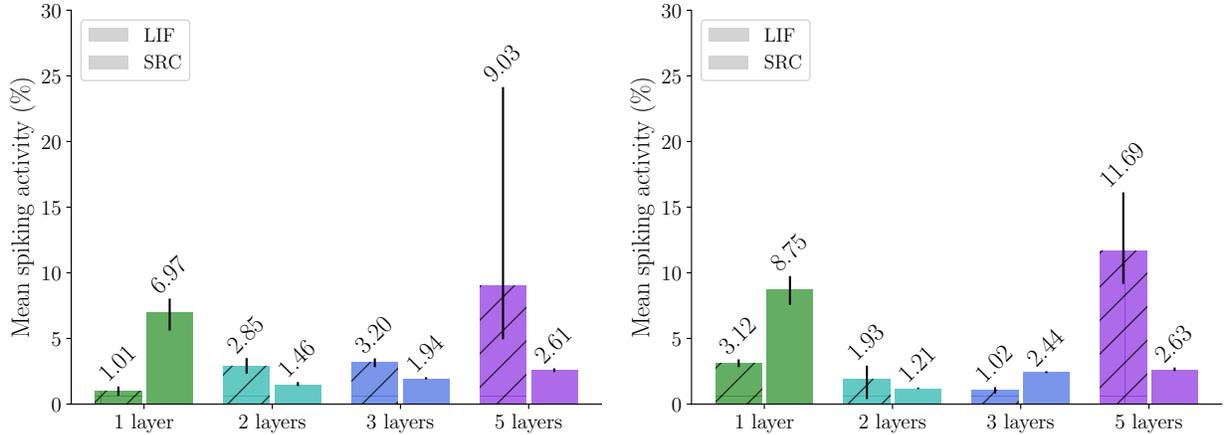

    \centering
    \includegraphics[width=.49\textwidth]{assets/pdf/from_wandb/spiking_rate.pdf}
    \includegraphics[width=.49\textwidth]{assets/pdf/from_wandb/spiking_latency.pdf}
    \caption{Mean spiking activity on the test set of the MNIST dataset with rate-based coding (\textbf{left}) and latency-based coding (\textbf{right}).}
    \label{fig:spiking}
\end{figure}

As previously mentioned, another important aspect of such networks is neuron activity.
Using the measure defined in \autoref{eq:spiking_time}, the mean activity of the neurons has been computed on the test set of the different benchmarks.
As a reminder, we have defined this activity as the relative number of timesteps where a spike is being emitted.
For the LIF, it corresponds to the relative number of spikes, while it is not the case for SRC.
\autoref{fig:spiking} reports these values for the MNIST dataset with the two codings.
The same tendency can be observed with both codings: SRC networks tend to spike more when there is no hidden layer, while it is the opposite for LIF networks.
In overall, the activity of both types of neurons is comparable.

\subsubsection{Training deeper neural networks}

Shallow networks of SRC neurons have successfully been trained.
However, one of important breakthroughs in deep learning was the ability to train deep neural networks.
Training deep SNNs is known to be difficult.
We have therefore tested networks with more layers to see if SRCs manage to learn also when the network becomes deeper.
These were made on the MNIST dataset with the rate-based coding and with 10 and 15 layers.

\begin{figure}[htbp]
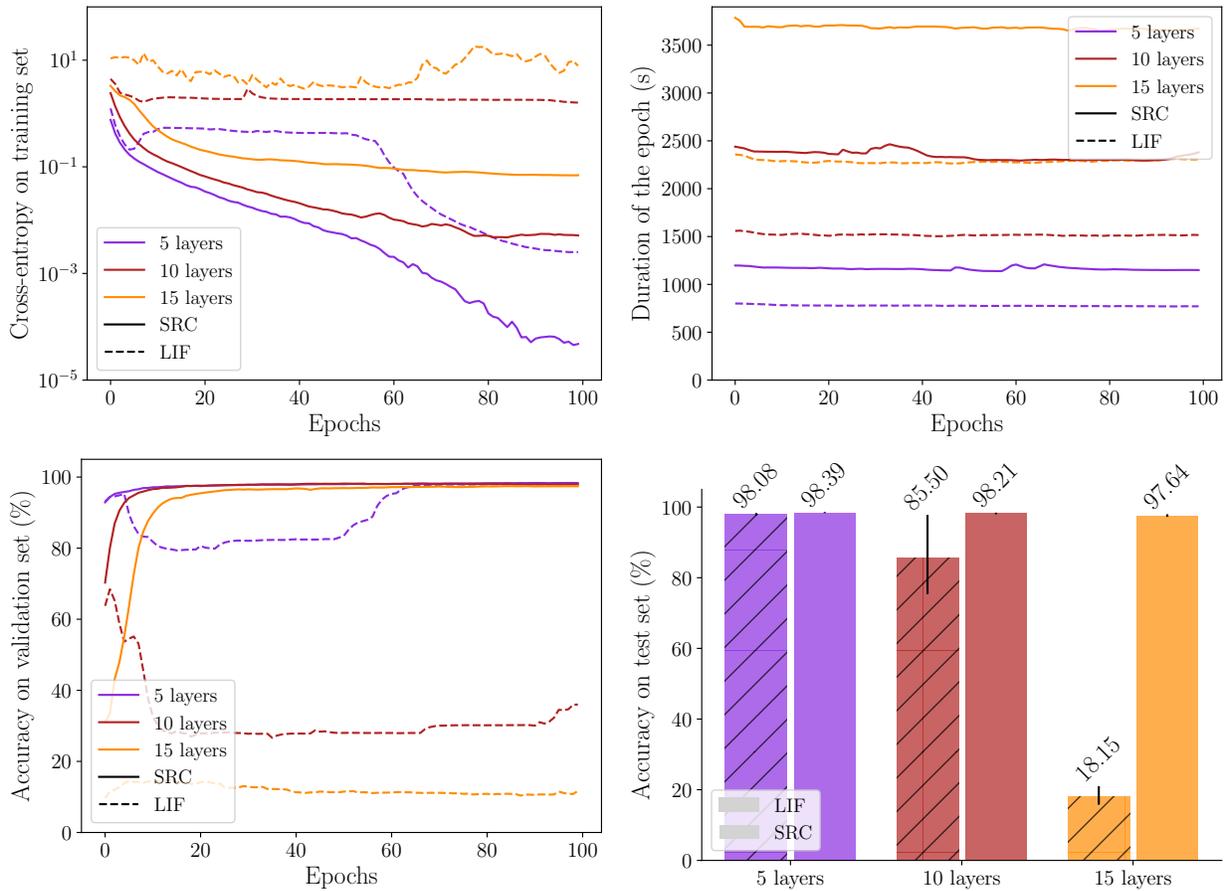

    \centering
    \begin{subfigure}{\textwidth}
        \centering
        \includegraphics[width=.49\textwidth]{assets/pdf/from_wandb/hl_ce.pdf}
        \includegraphics[width=.49\textwidth]{assets/pdf/from_wandb/hl_duration.pdf}
    \end{subfigure}
    \begin{subfigure}{\textwidth}
        \centering
        \includegraphics[width=.49\textwidth]{assets/pdf/from_wandb/hl_valid_acc.pdf}
        \includegraphics[width=.49\textwidth]{assets/pdf/from_wandb/hl_test_acc.pdf}
    \end{subfigure}
    \caption{(\textbf{left}) Evolution of the cross-entropy (\textbf{top}) and the accuracy (\textbf{bottom}) on the validation set during the training of several networks with 5, 10 and 15 layers. (\textbf{top, right}) Evolution of the epoch duration in seconds of these networks. (\textbf{bottom, right}) Cross-entropies and accuracies obtained on the MNIST test set by the models with respect to their number of hidden layers.}
    \label{fig:exp_hidden_layers}
\end{figure}

\autoref{fig:exp_hidden_layers} shows the results of this experiment.
To compare with more shallow networks, the results obtained with the 5-layers networks have been added.
From this figure we can observe that SRC leads to much more stable learning for a high number of layers.
Moreover, LIF networks completely fail to learn when they are composed of 15 layers, while for 10 layers they achieve honorable results but with a very unstable training.
Finally, as SRC is a more complex neuron than LIF, we wondered what was the difference in the training durations.
The top-right graph shows the mean duration of the epochs for each network.
As expected, SRC networks take more time, but still stay in the same order of magnitude as LIF networks.

\begin{new}
    \subsubsection{Introducing noise in SRC dynamic}

    \begin{figure}[htbp]
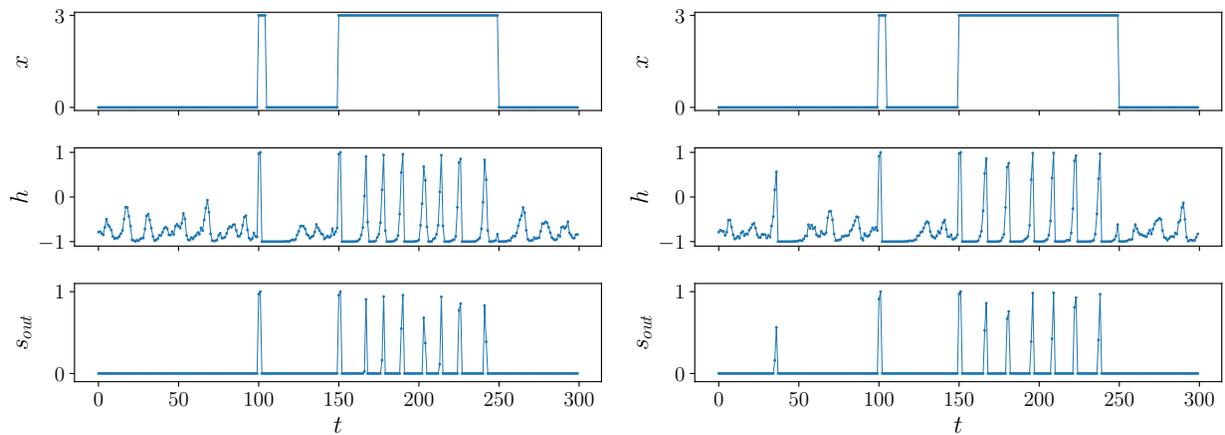

        \centering
        \includegraphics[width=.49\textwidth]{assets/pdf/src_dynamics/noise1.pdf}
        \includegraphics[width=.49\textwidth]{assets/pdf/src_dynamics/noise2.pdf}
        \caption{Two simulation of SRC with the same input stimulation and with $\sigma_{noise} = 0.2$}
        \label{fig:src_noise}
    \end{figure}

    One major difference between LIF and SRC is the shape of the spikes: the spikes of LIF neurons always last exactly one timestep and have a value of $1$, which ensures that no information can be transmitted through the spike shape.
    On the other hand, SRC spikes last several timesteps, and take continuous values between $0$ and $1$.
    One may wonder if trained SRC networks may learn to transmit information by \textit{modulating} the shape of the spikes.
    Indeed, despite having no control on the \textit{feedback} parameters ($r$ and $r_s$) and on the \textit{update} parameters ($z$ and $z_s$), the models may learn to shape the spike through the inputs.
    To check this, we implemented a slightly different version of SRC, where $r$ and $r_s$ are not fixed anymore, but sampled from two normal distributions:
    \begin{align*}
        r   & \sim \mathcal{N}(\mu_r, \sigma_{noise})     \\
        r_s & \sim \mathcal{N}(\mu_{r_s}, \sigma_{noise}) \\
    \end{align*}
    where $\mu_r$ and $\mu_{r_s}$ are the fixed values that we used previously, i.e. $2$ and $-7$, and $\sigma_{noise}$ is the level of noise we want to apply.
    \autoref{fig:src_noise} shows two simulations of SRC (with only the spike-generation part) where some noise was applied as explained before.
    It is clear that this added noise prevents from having the same output if the same input is received.

    \begin{figure}[htbp]
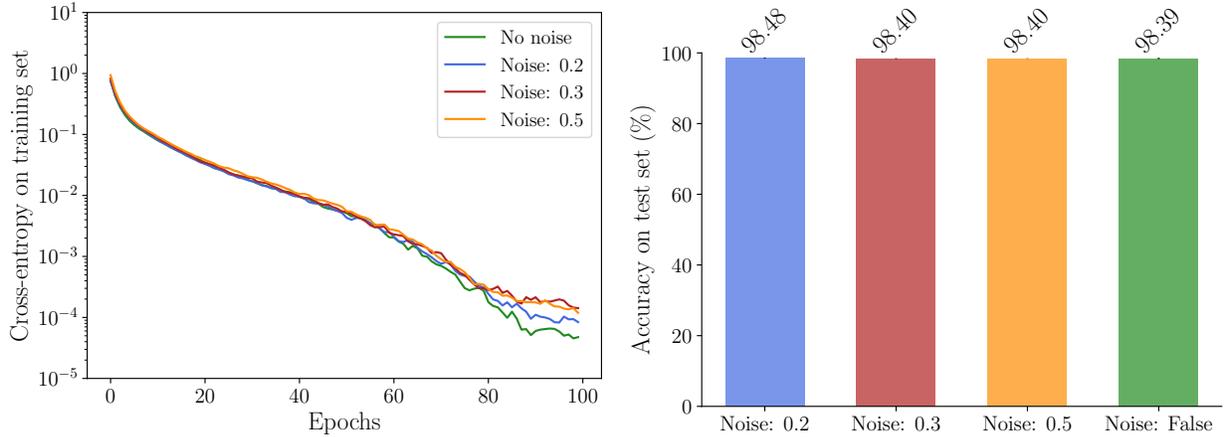

        \centering
        \includegraphics[width=.49\textwidth]{assets/pdf/from_wandb/noise_loss.pdf}
        \includegraphics[width=.49\textwidth]{assets/pdf/from_wandb/noise_acc.pdf}
        \caption{Evolution of the training cross-entropy (\textbf{left}) and test accuracy (\textbf{right}) of 5-layers SRC networks with several level of noises on the rate-based coded MNIST.}
        \label{fig:experiments_noise}
    \end{figure}

    \autoref{fig:experiments_noise} shows the results of experiments on the rate-coded MNIST, with SRC networks of 5 layers and several level of noises. Aside from the introduction of noise, all parameters are kept identical. For each level of noise, 5 models have been trained. The introduction of noise has nearly zero impact on the performance and the learning of the networks.
    It is therefore safe to say that the performance obtained by SRC networks does not rely on the \textit{non-binary} shape of its spikes.
\end{new}

\section{Conclusion}
\label{sec:conclusion}

In this paper, we have introduced a new type of artificial spiking neuron.
Instead of deriving this neuron from existing spiking models, as it is classically done, we have started from a largely used RNN cell.
This new spiking neuron, called the \textit{Spiking Recurrent Cell} (\textbf{SRC}), can be expressed as a usual recurrent cell.
Its major advantage is the differentiability of its equations.
This property allows to directly apply the usual backpropagation algorithm to train SRCs.
Spiking neural networks made of SRCs have been tested on the MNIST benchmark as well as two variants, the Fashion MNIST and the Neuromorphic MNIST.
These networks have achieved results which are comparable to the ones obtained with other non-convolutional SNNs.
Furthermore, training deep SNNs with LIF neurons is very fastidious and unstable, while using SRCs in such networks make the training much easier.
This proof of concept shows promising results and paves the way for new experiments.
On the machine learning side, it would be interesting to test SRC on more complex benchmarks to check its capabilities.
Like it is done with LIFs, it is possible to make a convolutional version of SRC, to use it on more difficult image classification tasks.
Also, adding feedback connections could increase the computational power of the SRC, as it is up to now only a feedforward SNN.
On the neuromorphic engineering side, it is feasible to improve SRC equations to make it exhibit new biological behaviors such as bursting for instance.
It would be very interesting to see if applying backpropagation to a spiking neuron that can have different firing patterns leads to an utilization of these different patterns depending on the context.
It is also possible to apply neuromodulation on the internal parameters of SRC in order to adapt neuron response is a context-dependent manner.

\begin{new}
Finally, SRC networks gain in energy efficiency can only be obtained when running on specialized, neuromorphic hardware.
Most neuromorphic chips created nowadays are designed to run LIF networks, and they may not be able to run SRCs.
The greatest difficulty lies in the shape of the spikes generated by SRC, which take continuous values, unlike LIF spikes.
First, it is likely that SRC does not require high precision real numbers, as we have seen that adding noise in the shape of the spikes does not affect performance. It should there be possible to quantize output values, which would make the implementation of SRC on such hardware much easier. Second, SRC being derived from GRU cells, one could implement an energy-efficient version on a FPGA, taking advantage of the work that has been done for GRU cells \citep{zaghloul2021FPGA,derick2018Simulation}.
Third, the continuous nature of SRC spikes could be exploited in fully-analog spiking systems \citep{chicca2014Neuromorphic}.
\end{new}

\section*{Acknowledgments}
Florent De Geeter gratefully acknowledges the financial support of the Walloon Region for Grant No. 2010235 – ARIAC by DW4AI.

\begin{new}
\section*{Data availability statement}
No new data were created or analysed in this study.
\end{new}

\bibliography{main.bib}

\newpage
\appendix

\section{Experiments details}
\label{sec:experiments_details}

This section gives the details about the experiments whose results are presented in \autoref{sec:results}. Note that, unless specified, the parameters of the neurons are not learnable.

\vspace{1em}

\begin{center}
    \begin{tabular}{|ll|}
        \hline
        \textbf{Name}                               & \textbf{Value}                                                      \\
        \hline
        \multicolumn{2}{|l|}{\textbf{For training}}                                                                       \\
        Number of epochs                            & $100$ (except on Fashion-MNIST: $200$)                              \\
        Batch size                                  & $64$                                                                \\
        Learning rate                               & Cosine annealing starting at $0.005$                                \\
        Optimizer                                   & Adam                                                                \\
        Clipping gradients norm                     & $1.$                                                                \\
        Ratio of training and validation sets sizes & $90\%/10\%$                                                         \\
        \hline
        \multicolumn{2}{|l|}{\textbf{For datasets}}                                                                       \\
        Sequences length                            & $200$                                                               \\
        Gain (for rate-based coding)                & $0.25$                                                              \\
        $\tau$ (for latency-based coding)           & $10$                                                                \\
        $V_{th}$ (for latency-based coding)         & $0.01$                                                              \\
        \hline
        \multicolumn{2}{|l|}{\textbf{For networks}}                                                                       \\
        Size of hidden layers                       & $512$                                                               \\
        Size of final layer                         & $10 \, \times$ the number of outputs                                \\
        Excitatory weight of readout                & $1.$                                                                \\
        Inhibitory weight of readout                & $-0.1$                                                              \\
        $\alpha$ of readout integrators             & $0.99$                                                              \\
        Initialization of weights                   & Xavier uniform                                                      \\
        \hline
        \multicolumn{2}{|l|}{\textbf{For SRC}}                                                                            \\
        $\alpha$                                    & $0.9$ (except for latency-based coding: $0.99$)                     \\
        $\rho$                                      & $3.$                                                                \\
        $z$                                         & $0.$                                                                \\
        $z_s^{hyp}$                                 & $0.9$                                                               \\
        $z_s^{dep}$                                 & $0.$                                                                \\
        $r$                                         & $2.$                                                                \\
        $r_s$                                       & $-7.$                                                               \\
        $b_h$                                       & $-6.$, \textbf{learnable}                                           \\
        max $b_h$                                   & $-4.$                                                               \\
        \hline
        \multicolumn{2}{|l|}{\textbf{For LIF}}                                                                            \\
        $\alpha_V$                                  & $0.9$ (except for latency-based coding: $0.99$), \textbf{learnable} \\
        $V_{thresh}$                                & $1.$, \textbf{learnable}                                            \\
        $V_{rest}$                                  & $0.$                                                                \\
        Surrogate activation                        & ATan                                                                \\
        \hline
    \end{tabular}
\end{center}

\newpage

\section{Feedback dynamics in SRC}
\label{sec:feedback_dynamics_in_src}

This section analyses the interaction of the two feedbacks $r h$ and $r_s h_s$ that allow to generate spikes in a SRC.

\begin{figure}[htb]
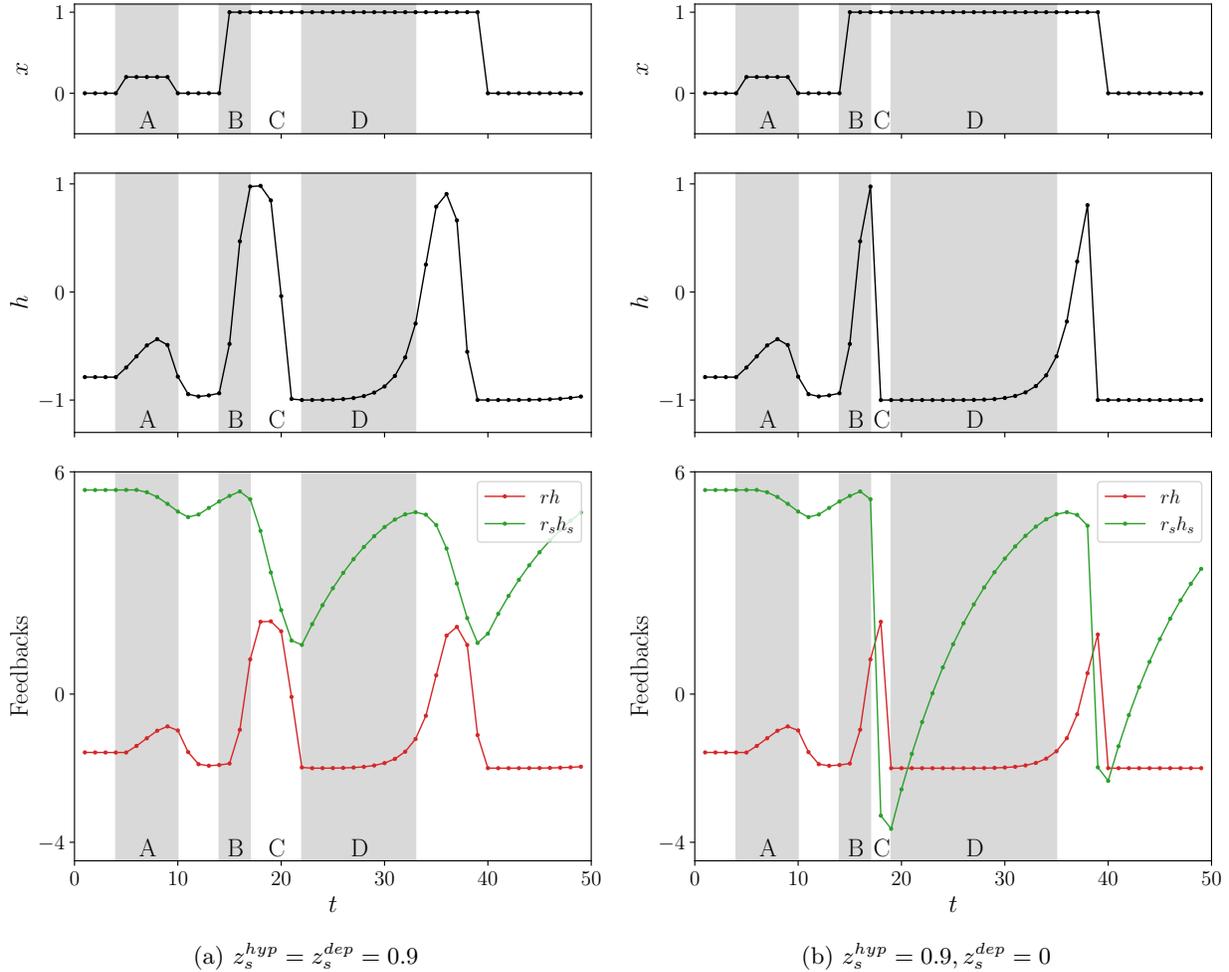

    \centering
    \begin{subfigure}{.49\textwidth}
        \centering
        \includegraphics[width=\textwidth]{assets/pdf/src_dynamics/spike_fix_zs.pdf}
        \caption{$z_s^{hyp} = z_s^{dep} = 0.9$}
        \label{fig:feedbacks_src_fix}
    \end{subfigure}
    \begin{subfigure}{.49\textwidth}
        \centering
        \includegraphics[width=\textwidth]{assets/pdf/src_dynamics/spike.pdf}
        \caption{$z_s^{hyp} = 0.9, z_s^{dep} = 0$}
        \label{fig:feedbacks_src_var}
    \end{subfigure}
    \caption{Evolution of $h$ and the feedbacks $r h$ (positive) and $r_s h_s$ (negative) given a stimulation of inputs $x$. Two versions of SRC are compared: when $z_s$ is fixed, and when it is not. Four parts of the simulation are highlighted. \textbf{A} The stimulation is not high enough, the threshold is not crossed and no spike is emitted. The negative feedback quickly overtakes the positive feedback. \textbf{B} The stimulation is high enough, the threshold is crossed and the positive feedback pushes $h$ towards higher values. \textbf{C} The negative feedback overtakes the positive feedback, creating the descending side of the spike. When $z_s$ is variable, the negative feedback is much quicker at this particular moment, as $z_s$ switches to $z_s^{dep}$. \textbf{D} The spike is finished but the negative feedback still applies, preventing another spike to happen. This can be seen as the \textit{refactory} period. When $z_s$ is variable, it switches back to $z_s^{hyp}$ making it much slower. Once this period is finished, another spike is emitted.}
    \label{fig:feedbacks_src}
\end{figure}

\autoref{fig:feedbacks_src} shows a simulation of the spike generation part of SRC when some stimulation is applied.
Two versions of SRC are compared: whether $z_s$ is fixed or not.
The advantage of the variable $z_s$ is twofolds.
First, the duration of the spike is smaller, thanks to a very fast descending side.
Then, during the refactory period, $z_s$ switches back to $z_s^{hyp}$, making the convergence of $h_s$ towards $h$ much slower.
This decoupling between the convergence speed of $h_s$ during the spike and during the refactory period allows to control the duration of this period without any impact on the spike duration.

\end{document}